\documentclass{article}
\usepackage{arxiv_ml_institute}

\usepackage[utf8]{inputenc} 
\usepackage[T1]{fontenc}    
\usepackage{hyperref}       
\usepackage{url}            
\usepackage{booktabs}       
\usepackage{amsfonts}       
\usepackage[nice]{nicefrac} 
\usepackage{microtype}      
\usepackage{xcolor}         
\usepackage{graphicx}       

\usepackage{amsmath,amsthm,amssymb,bm} 
\usepackage{xspace}
\usepackage{array}
\usepackage{enumerate}
\usepackage{enumitem}
\usepackage{stmaryrd}
\usepackage{caption}
\usepackage{subcaption}
\usepackage{multirow}
\usepackage{times}
\usepackage{natbib}
\usepackage{algorithm}
\usepackage{algorithmicx,algpseudocode}
\usepackage{hyperref}
\usepackage{sidecap}
\usepackage{booktabs}
\usepackage{wrapfig}
\usepackage{fancyhdr}

\hypersetup{
   colorlinks=true,
   linkcolor=JKUblue,
   citecolor=JKUblue,
   urlcolor=magenta,
   pdfborder=0 0 0,
   pdftitle={},
   pdfsubject={}, 
   pdfkeywords={},
   pdfauthor={},
   pdfstartview=FitH
}

\usepackage[mathscr]{eucal}
\usepackage{pifont}
\usepackage{cleveref}

\usepackage[most]{tcolorbox}
\newtcolorbox[auto counter]{NoteBox}[2][]{
  title={\textbf{Note:} #2 }, 
  boxsep=2mm,
  coltitle=black,
  colframe=gray!35,
  colback=gray!7,
  breakable,
  #1
}

\definecolor{JKUblue}{RGB}{0, 132, 187}
\definecolor{JKUviolet}{RGB}{148,0,187}
\definecolor{JKUred}{RGB}{187,55,0}
\definecolor{JKUgreen}{RGB}{39,187,0}

\newtheorem{lemma}{Lemma}%
\newcommand\Bh{\bm{h}}

\newcommand\Bm{\bm{m}}

\newcommand\Bw{\bm{w}}

\newcommand\BI{\bm{I}}

\newcommand\Bmu{\bm{\mu}}

 \newcommand{\dR}{\mathbb{R}}

\newcommand\EXP{\mathbf{\mathrm{E}}}

\newcommand\VAR{\mathbf{\mathrm{Var}}}
\newcommand\COV{\mathbf{\mathrm{Cov}}}

\usepackage{bigdelim}
\usepackage{colortbl} 
\definecolor{mColor1}{rgb}{0.95,0.95,0.95}

\Crefname{equation}{Eq.}{Eqs.}
\Crefname{figure}{Fig.}{Figs.}
\Crefname{tabular}{Tab.}{Tabs.}
\Crefname{appendix}{App.}{Apps.}

\newcommand{\mycheck}{\raisebox{0.0ex}{\color{green} \scriptsize \ding{52} } }
\newcommand{\myfail}{\raisebox{0.0ex}{\color{red} \scriptsize \ding{56} }}
\newcommand{\myunkn}{\raisebox{0.0ex}{\color{orange} \scriptsize \ding{109} }}

\title{GNN-VPA: A Variance-Preserving Aggregation Strategy for Graph Neural Networks}

\author{
  Lisa Schneckenreiter$^{1}$\thanks{Equal contribution}
  \ \ \ \ \ \ \ \ \ \ \ 
  Richard Freinschlag$^{1}$\footnotemark[1]
  \ \ \ \ \ \ \ \ \ \ \ 
  Florian Sestak$^{1}$
  \ \ \ \ \ \ \ \ \ \ \ \\ \\
  \textbf{Johannes Brandstetter}$^{1,2}$
  \ \ \ \ \ \ \ \ \ \ \
  \textbf{Günter Klambauer}$^{1}$
  \ \ \ \ \ \ \ \ \ \ \
  \textbf{Andreas Mayr}$^{1}$
  \ \ \ \ \ \ \ \ \ \ \ \\
  \\ \\
{$^1$~ELLIS Unit Linz and LIT AI Lab, Institute for Machine Learning,} \\{Johannes Kepler University, Linz, Austria}\\
{$^2$~NXAI GmbH, Linz, Austria}\\
  \texttt{last-name@ml.jku.at}
}

\fancypagestyle{empty}{
  \fancyhf{}
  \fancyhead[L]{Accepted at ICLR 2024 (Tiny Papers Track)}
  \fancyfoot[C]{\thepage}
}

\begin{document}

\pagestyle{fancy}
\fancyhf{}
\fancyhead[L]{Accepted at ICLR 2024 (Tiny Papers Track)}
\fancyfoot[C]{\thepage}

\maketitle

\begin{abstract}
    Graph neural networks (GNNs), and especially message-passing neural networks, excel in various domains such as physics, drug discovery, and molecular modeling. The expressivity of GNNs with respect to their ability to discriminate non-isomorphic graphs critically depends on the functions employed for message aggregation and graph-level readout. By applying signal propagation theory, we propose a variance-preserving aggregation function (VPA) that maintains expressivity, but yields improved forward and backward dynamics. Experiments demonstrate that VPA leads to increased predictive performance for popular GNN architectures as well as improved learning dynamics. Our results could pave the way towards normalizer-free or self-normalizing GNNs.
\end{abstract}

\section{Introduction and related work}

For many real-world prediction tasks, graphs naturally represent the input data. Graph neural networks (GNNs) \citep{Scarselli2009, kipf2016semi, defferrard2016convolutional, Velickovic:18} are therefore of large interest as they are able to naturally process such data. They have been used for molecule predictions \citep{Duvenaud2015, Kearnes2016, gilmer2017neural, Mayr2018, Satorras2021}, material science \citep{Reiser2022, Merchant2023}, modeling physical interactions or improving PDE solvers for physics predictions \citep{Sanchez2020, Brandstetter2022b, Mayr2023}, weather prediction \citep{keisler2022forecasting, lam2022graphcast}, predictions about social networks \citep{Hamilton2017, Fan2019, Monti2019}, gene regulatory networks in systems biology \citep{Eetemadi2018, Wang2020}, combinatorial optimization \citep{Cappart2023, Sanokowski2023}, and knowledge graphs \citep{Schlichtkrull2018, Li2022} for reasoning.

Despite the huge successes of GNNs, there are some limitations. \citet{Morris2019} and \citet{Xu2019} analyzed the expressive power of GNNs and found that they are not more powerful than the Weisfeiler-Leman graph isomorphism heuristic (1-WL test) \citep{Leman1968} at distinguishing non-isomorphic graphs. Moreover, \citet{Xu2019} constructed a GNN (\textsc{GIN} architecture), which should attain the same expressive power as the 1-WL test.
An important conclusion in the design of the \textsc{GIN} architecture was that the choice of the message aggregation and graph-level readout function is crucial for enabling maximum expressivity. More specifically, \textsc{sum} aggregation allows to attain 1-WL expressive power, while \textsc{mean} or \textsc{max} aggregation effectively limits expressivity.

While the expressive power of GNNs has been investigated profoundly \citep{Xu2019}, signal propagation \citep{neal1995bayesian,schoenholz2016deep,klambauer2017self} through GNNs is
currently under-explored. There are plenty of works on conventional fully-connected neural networks (FCNNs), which study signal propagation behavior \citep[e.g.,][]{schoenholz2016deep, klambauer2017self} throughout the networks. Typically, for FCNNs or convolutional neural networks (CNNs), there are either weight initialization schemes \citep[e.g.,][]{Glorot2010, He2015} or normalization layers \citep[e.g.,][]{Ioffe2015, Ba2016}, which prevent that the weighted summed inputs lead to exploding activations throughout the depth of the network. 

For GNNs and especially the \textsc{GIN} architecture with \textsc{sum} message aggregation, exploding activations are a main obstacle for efficient training as well and signal propagation behavior appears problematic. Conventional weight initialization schemes at the aggregation step cannot be applied in a straightforward manner, since the number of neighbors in an aggregation step and the number of nodes in a graph are variable. Moreover, the fact that zero variance in messages might be a common case for graph classification also limits the applicability of normalization layers. 

\begin{figure}[h]
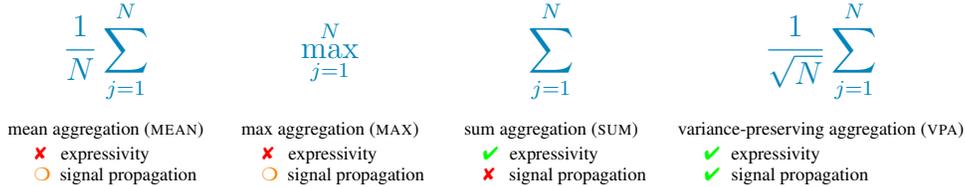

\begin{center}
\renewcommand{\arraystretch}{0.7}
\begin{tabular}{llll}
\multicolumn{1}{c}{\color{JKUblue}\large $\displaystyle \frac{1}{N} \sum_{j=1}^N$} & 
\multicolumn{1}{c}{\color{JKUblue} \large $\displaystyle \max_{j=1}^N$} & 
\multicolumn{1}{c}{\color{JKUblue} \large $\displaystyle \sum_{j=1}^N$} & 
\multicolumn{1}{c}{\color{JKUblue} \large $\displaystyle \frac{1}{\sqrt{N}} \sum_{j=1}^N$}\\
\\
\multicolumn{1}{c}{\raisebox{0.4ex}{\scriptsize mean aggregation (\textsc{mean})}} &
\multicolumn{1}{c}{\raisebox{0.4ex}{\scriptsize max aggregation (\textsc{max})}} &
\multicolumn{1}{c}{\raisebox{0.4ex}{\scriptsize sum aggregation (\textsc{sum})}} & 
\multicolumn{1}{c}{\raisebox{0.4ex}{\scriptsize variance-preserving aggregation (\textsc{vpa})}} \\
\quad \myfail \scriptsize  \text{ } expressivity & \quad \myfail \scriptsize \text{ }  expressivity &   \quad \mycheck \scriptsize expressivity & \quad \mycheck \scriptsize expressivity\\
\quad \myunkn \scriptsize \text{ }signal propagation & \quad \myunkn \scriptsize \text{ }signal propagation & \quad \myfail \scriptsize \text{ } signal propagation & \quad \mycheck \scriptsize  signal propagation\\
\end{tabular}
\end{center}
\caption{Overview of main message aggregation functions and their properties.}
\label{fig:enter-label}
\end{figure}

Our aim in this work is to develop a general aggregation approach \footnote{We are not interested in proposing a new pooling mechanism, but in suggesting a new aggregation function that can optionally be applied to graph-level readout. For further details on the differences between aggregation and pooling, see \Cref{sec:agg_vs_pool}.}, which can be applied to different GNN architectures, preserves maximum expressivity, and at the same time avoids exploding activations. With simplistic assumptions, we will motivate the use of a variance-preserving aggregation function for GNNs (see \Cref{fig:enter-label}), which improves signal propagation and consequently learning dynamics.

\section{GNNs with Variance Preservation}

\textbf{Notational preliminaries. } We assume a graph $\mathscr G= (\mathscr V,\mathscr E)$ with nodes $v_i \in \mathscr V$, edges $e_{ij} \in \mathscr E$ and $D$-dimensional node features $\Bh_i \in \dR^{D}$. We use $\mathscr N(i)$ to indicate the set of neighboring 
nodes to node $v_i$ within $\mathscr V$. To be consistent with \Cref{fig:enter-label}, we define $N$ always to be the number of neighboring 
nodes, i.e. $N:=|\mathscr N(i)|$, where we assume that $i$ is clear from the context. For simplicity, we do not assume any edge features.

\textbf{Graph neural networks} (GNNs)
exchange information, i.e., messages, through the application of a local, permutation-invariant function across all neighborhoods.
The core layers iteratively update node embeddings $\Bh_i$ at node $v_i$ via three substeps 1.-3.:
\begin{align*}
\text{1.\, }\Bm_{ij}=\phi\Big(\Bh_i, \Bh_j \Big) \ & \text{\, or \,} \ \Bm_{ij}=\phi\Big( \Bh_j \Big) &
\text{2.\, }\Bm^\oplus_{i} &= \bigoplus_{j \in {\mathscr N(i)}} \ \Bm_{ij} &
\text{3.\, }\Bh'_i &= \psi\Big(\Bh_i, \theta \Big( \Bm^\oplus_{i}\Big) \Big) \label{eq:message} 
\end{align*}

\noindent to a new embedding $\Bh'_i$, where the aggregation $\bigoplus_{j \in {\mathscr N(i)}}$ at node $v_i$ is across all neighboring nodes, i.e., those nodes $v_j$, that are connected to node $v_i$ via an edge $e_{ij}$. These nodes are renumbered according to \Cref{fig:enter-label} from $1$ to $N$. Depending on the type of GNN, $\phi$, $\psi$, and $\theta$ can be realized as learnable functions, usually Multilayer Perceptrons (MLPs). E.g., for Graph Convolutional Networks (GCNs) \citep{kipf2016semi} only 
$\psi$ is learnable, for general Message Passing Neural Networks \citep{gilmer2017neural} $\phi$ and $\psi$ are learnable, and for Graph Isomorphism Networks (GINs) \citep{Xu2019} all three are learnable.

\textbf{Signal propagation theory} allows to analyze
the distribution of quantities through randomly initialized neural networks. From certain assumptions (for details see \Cref{sec:signal_prop_mpnn}) it follows that $\Bm_{ij} \sim p_{\mathcal N}(\boldsymbol 0, \BI)$. If $\Bm_{ij}$ are further assumed to be independent of each other\footnote{Note 
that this assumption is too strong, since for a fixed $i$, 
all $\Bm_{ij}$ depend on each other because they are all determined by the input $\Bh_i$.}, one obtains $\Bm^\text{\scalebox{0.7}{\textsc{sum}}}_{i} \sim p_{\mathcal N}(\boldsymbol 0, N  \BI)$ for \textsc{sum} aggregation (i.e., $\bigoplus \equiv \sum_{i=1}^N$), and $\Bm^\text{\scalebox{0.7}{\textsc{mean}}}_{i} \sim p_{\mathcal N}(\boldsymbol 0, \frac{1}{N}\BI)$ for \textsc{mean} aggregation (i.e., $\bigoplus \equiv \frac{1}{N}\sum_{i=1}^N$) at initialization.

\textbf{A variance preserving aggregation (VPA) function. } Our key idea is to introduce a new aggregation function which preserves variance, i.e.,  $\Bm_{i} \sim p_{\mathcal N}(\boldsymbol 0, \BI)$. This is possible with the aggregation function $\bigoplus \equiv \frac{1}{\sqrt{N}} \sum_{i = 1}^N$. 
We denote this aggregation function as \emph{variance-preserving aggregation} (\textsc{vpa}) and show the preservation property
by applying {\color{JKUblue} Lemma}~\ref{lemma:vpp}
element-wise. For a complete proof see App. \ref{sec:var_pres}.

\begin{lemma} 
\label{lemma:vpp}
Let $z_1,\ldots, z_N$ be independent copies of a centered random 
variable $z$ with finite variance. Then the random variable 
$y=\frac{1}{\sqrt{N}}\sum_{n=1}^N z_n$ has the same mean and variance as $z$.
\end{lemma}

In contrast to \textsc{sum} or \textsc{mean} aggregation functions, \textsc{vpa} theoretically preserves the variance across layers.
According to signal propagation theory, such behavior is advantageous for learning.

\textbf{Expressive power of GNN-VPA. } According to \citet{Xu2019} a prerequisite for maximum expressive power w.r.t. discriminating non-isomorphic graphs is an injective aggregation function, such as \textsc{sum} aggregation, while \textsc{mean} or \textsc{max} aggregation results in limited expressivity. A message passing algorithm with \textsc{vpa} has the same expressive power as \textsc{sum} aggregation, which follows analogously to \citet{Xu2019} from {\color{JKUblue} Lemma}~\ref{lemma:expressivity} (see \Cref{sec:expr} for a proof).

\begin{lemma}
    \label{lemma:expressivity}
    Assume the multiset $\mathcal{X}$ is countable and the number of unique elements in $\mathcal{X}$ is bounded by a number $N$. There exists a function $f: \mathcal{X} \rightarrow \mathbb{R}^N$ such that $h(X) = \frac{1}{\sqrt{|X|}} \sum_{x \in X} f(x)$ is unique for each multiset $X \subset \mathcal{X}$ of bounded size, where $|X|$ denotes the cardinality of multiset $X$ (sum of multiplicities of all unique elements in the multiset).
\end{lemma}

\textbf{Extension of variance preservation to attention. } Our variance-preserving aggregation strategy can be extended to attention mechanisms. We assume, that random variables $z_1,\ldots, z_N$ are aggregated by an attention mechanism. The respective computed attention weights are assumed to be given by $c_1,\ldots, c_N$, where $c_i \in \mathbb{R}_0^+$ and $\sum_{i=1}^N c_i=1$ holds. Further, we consider $c_i$ to be constants \footnote{Note, that this might be an over-simplistic assumption, especially since/when keys and values are not independent.}.

In order to find a useful extension of \textsc{vpa} to attention, we first consider two extreme cases on the distribution of attention weights:
\begin{itemize}
    \item Case 1: All attention weights are equal. Then in order to fulfill $\sum_{i=1}^N c_i=1$, all $c_i=\frac{1}{N}$.
    \item Case 2: Attention focuses on exactly one value, which might be w.l.o.g. $j$. Then $c_j=1$ and $c_i=0 \ \ \forall \ i \neq j$.
\end{itemize}

We note that case 1 is the same as \textsc{mean} aggregation and case 2 corresponds to \textsc{max} aggregation if $max(z_1,\ldots, z_N)=z_j$ and $z_i<z_j  \ \ \forall \ i \neq j$. In both cases, GNNs have more limited expressivity than with \textsc{vpa} or \textsc{sum} aggregation.

To apply the concept of variance preservation to attention, we define a constant $C:=\sqrt{\sum_{i=1}^N c_i^2}$ and use the following attention mechanism: $y=\frac{1}{C} \ \sum_{i=1}^N c_i \ z_i$. As shown in {\color{JKUblue} Lemma}~\ref{lemma:vpp_att} this results in a variance-preserving attention mechanism. For a complete proof see \Cref{sec:proof_gat}. 

\begin{lemma} 
\label{lemma:vpp_att}
Let $z_1,\ldots, z_N$ be independent copies of a centered random 
variable $z$ with finite variance and let $c_1,\ldots, c_N$ be constants, where $c_i \in \mathbb{R}_0^+$ and $\sum_{i=1}^N c_i=1$. Then the random variable 
$y=\frac{1}{C}\sum_{n=1}^N c_n z_n$ with $C=\sqrt{\sum_{i=1}^N c_i^2}$ has the same mean and variance as $z$.
\end{lemma}

\section{Experiments}
\label{sec:experiments}

We tested the effectiveness of our idea on a range of established GNN architectures\footnote{Code is available at \url{https://github.com/ml-jku/GNN-VPA}.}: Graph Isomorphism Networks (\textsc{GIN}) \citep{Xu2019}, Graph Convolutional Network (\textsc{GCN}) \citep{kipf2016semi}, Graph Attention Networks (\textsc{GAT}) \citep{Velickovic:18} and Simple Graph Convolution Networks (\textsc{SGC}) \citep{sgc}. To evaluate prediction performance, we combined \textsc{GIN} and \textsc{GCN} architectures with each of the aggregation methods in \Cref{fig:enter-label} both for message aggregation and graph-level readout. Note that we used the \textsc{GCN} formulation as reported in \citet{Morris2019} to circumvent the inherent normalization in the \textsc{GCN} architecture by \citet{kipf2016semi}. 

To incorporate the idea of variance preservation into the \textsc{SGC} architecture, we changed the update of $\mathbf{h}$ from $$\mathbf{h}'_i = \frac{1}{d_i + 1} \mathbf{h}_i+\sum_{j=1}^N \frac{a_{ij}}{\sqrt{(d_i + 1) (d_j + 1)}}\mathbf{h}_j$$ to $$\mathbf{h}'_i = \frac{1}{\sqrt{d_i + 1}} \mathbf{h}_i+\sum_{j=1}^N \frac{a_{ij}}{\sqrt[4]{(d_i + 1) (d_j + 1)}}\mathbf{h}_j$$
(where $a_{ij}$ are entries of the adjacency matrix, $d_i$ and $d_j$ are node degrees, and, $\mathbf{h_i}$ and $\mathbf{h_j}$ denote the hidden neural representation at some time step during message passing). For a variance-preserving version of \textsc{GAT}, we adapted attention according to 
~\Cref{lemma:vpp_att} and note that in the practical implementation, we do not backpropagate errors through these constants during training.

\textbf{Benchmarking datasets and settings.} We tested our methods on the same graph classification benchmarks from the TUDataset collection as \citet{Xu2019}, consisting of five social network datasets (IMDB-BINARY, IMDB-MULTI, COLLAB, REDDIT-BINARY, and REDDIT-MULTI-5K) and four bioinformatics datasets (MUTAG, PROTEINS, PTC and NCI1). Since the social network datasets do not contain any node features, we introduced node features in two different ways. In the first variant, the graphs are considered as given with all node features set to $1$. In the other variant, the one-hot encoded node degree is used as an additional node feature. We report results for the first variant in \Cref{tab:main_results} and results for the second variant in \Cref{tab:results2}. The bioinformatics datasets were used with the provided node features. For more details on the used datasets, we refer to \citet{Morris2020} and \citet{Xu2019}.

\textbf{Training, validation, and test splits.} Our experiments were evaluated with 10-fold cross-validation. In each iteration, we used $\nicefrac{1}{10}$ of the data for testing, $\nicefrac{1}{10}$ for validation and $\nicefrac{8}{10}$ for training. The validation set was only used to adjust the number of training epochs, such that our test accuracies were computed for the epoch with the highest validation accuracy. For more details on implementation and hyperparameters see \Cref{sec:hyperparams}.

\textbf{Results.} Test accuracies for all four GNN architectures comparing \textsc{vpa} with the standard aggregation methods are shown in ~\Cref{tab:main_results}. In almost all cases, \textsc{vpa} significantly outperforms the compared methods. Notably, the \textsc{GIN} and \textsc{GCN} architectures in combination with \textsc{mean} or \textsc{max} aggregation were unable to learn the social network tasks without additional node features, likely due to the inherent inability of these aggregation functions to capture a node's degree. This emphasizes the increased expressivity of  \textsc{vpa} compared to these methods. For additional results concerning the training behavior, see~\Cref{sec:learningCurves}.

\begin{table}[ht]
    \centering
    \resizebox{0.99\textwidth}{!}{
    \begin{tabular}{l|ccccccccc|cc}
        \toprule 
        &  {\textsc{IMDB-B}} & {\textsc{IMDB-M}}  & {\textsc{RDT-B}} & {\textsc{RDT-M5K}} & {\textsc{COLLAB}} & {\textsc{MUTAG}} &  {\textsc{PROTEINS}}  & {\textsc{PTC}} & {\textsc{NCI1}} & {\textsc{Avg}} & p\\ \midrule
        {\textsc{GIN+sum}} & 71.8  $\pm$ 4.0 & 47.1  $\pm$ 4.3 & 85.5  $\pm$ 2.2 & 52.0  $\pm$ 3.0 & 70.9  $\pm$ 1.5 & \bf 87.2  $\pm$ 4.9 & \bf 73.3  $\pm$ 3.1 & 54.1  $\pm$ 7.1 & \bf 81.7  $\pm$ 2.3 &  69.3 & 2.0e-5  \\ 
        {\textsc{GIN+mean}} & 50.0  $\pm$ 0.0 & 33.3  $\pm$ 0.0 & 50.0  $\pm$ 0.1 & 20.0  $\pm$ 0.1 & 32.5  $\pm$ 0.1 & 76.1  $\pm$ 11.1 & 67.2  $\pm$ 2.9 & 58.7  $\pm$ 6.5 & 77.7  $\pm$ 1.9 &  51.7 & 3.2e-15  \\ 
        {\textsc{GIN+max}} & 50.0  $\pm$ 0.0 & 33.3  $\pm$ 0.0 & 49.7  $\pm$ 0.5 & 20.2  $\pm$ 0.4 & 52.0  $\pm$ 0.0 & 77.0  $\pm$ 8.2 & 71.8  $\pm$ 3.6 & 59.0  $\pm$ 9.7 & 80.5  $\pm$ 2.8 &  54.8 & 3.9e-13  \\ 
        {\textsc{GIN+vpa}} & \bf 72.0  $\pm$ 4.4 & \bf 48.7  $\pm$ 5.2 & \bf 89.0  $\pm$ 1.9 & \bf 56.1  $\pm$ 3.0 & \bf 73.5  $\pm$ 1.5 & 86.7  $\pm$ 4.4 & 73.2  $\pm$ 4.8 & \bf 60.1  $\pm$ 5.8 & 81.2  $\pm$ 2.1 & \bf  71.2 &- \\ 
        \midrule
        {\textsc{GCN+sum}} & 63.3 $\pm$ 6.1 & 42.1 $\pm$ 3.7 & 75.4 $\pm$ 3.2 & 37.3 $\pm$ 3.5 & 67.0 $\pm$ 2.2 & \bf 78.7 $\pm$ 7.8 & 70.3 $\pm$ 2.2 & 61.3 $\pm$ 7.8 & \bf 80.2 $\pm$ 2.0 & 64.0 & 9.9e-9 \\ 
        {\textsc{GCN+mean}} & 50.0 $\pm$ 0.0 & 33.3 $\pm$ 0.0 & 49.9 $\pm$ 0.2 & 20.1 $\pm$ 0.1 & 52.0 $\pm$ 0.0 & 72.4 $\pm$ 6.3 & \bf 74.3 $\pm$ 4.4 & \bf 63.3 $\pm$ 6.5 & 75.8 $\pm$ 2.6 & 54.6 & 3.3e-12 \\
        {\textsc{GCN+max}} & 50.5 $\pm$ 0.0 & 33.3 $\pm$ 0.0 & 50.0 $\pm$ 0.0 & 20.0 $\pm$ 0.1 & 52.0 $\pm$ 0.0 & 67.6 $\pm$ 4.3 & 43.9 $\pm$ 7.3 & 58.7 $\pm$ 6.6 & 55.1 $\pm$ 2.6 & 47.8 & 1.4e-15 \\ 
        {\textsc{GCN+vpa}} & \bf 71.7 $\pm$ 3.9 & \bf 46.7 $\pm$ 3.5 & \bf 85.5 $\pm$ 2.3 & \bf 54.8 $\pm$ 2.4 & \bf 73.7 $\pm$ 1.7 & 76.1 $\pm$ 9.6 & 73.9 $\pm$ 4.8 & 61.3 $\pm$ 5.9 & 79.0 $\pm$ 1.8 & \bf 69.2 & -\\
        \midrule
        {\textsc{SGC}} & 62.9 $\pm$ 3.9 & 40.3 $\pm$ 4.1 & 78.9 $\pm$ 2.0 & 41.3 $\pm$ 3.5 & 68.0 $\pm$ 2.2 & 73.5 $\pm$ 9.8 & 73.1 $\pm$ 3.4 & 59.0 $\pm$ 6.0 & 68.5 $\pm$ 2.2 & 62.8 & 3.8e-12\\
        {\textsc{SGC+vpa}} & \bf 70.4 $\pm$ 4.1 & \bf 47.5 $\pm$ 4.4 & \bf 84.2 $\pm$ 2.2 & \bf 53.4 $\pm$ 2.7 & \bf 71.7 $\pm$ 1.7 & \bf 73.9 $\pm$ 6.2 & \bf 75.4 $\pm$ 4.2 & \bf 63.1 $\pm$ 8.0 & \bf 76.4 $\pm$ 2.8 & \bf 68.4 & -\\
        \midrule
        {\textsc{GAT}} & 51.0 $\pm$ 4.4 & 37.4 $\pm$ 3.6 & 74.5 $\pm$ 3.8 & 33.1 $\pm$ 1.9 & 56.2 $\pm$ 0.6 & 77.7 $\pm$ 11.5 & \bf 75.4 $\pm$ 2.9 & 60.5 $\pm$ 5.5 & \bf 77.7 $\pm$ 2.2 & 60.4 & 7.6e-9\\
        {\textsc{GAT+vpa}} & \bf 71.1 $\pm$ 4.6 & \bf 44.1 $\pm$ 4.5 & \bf78.1 $\pm$ 3.7 & \bf 43.3 $\pm$ 2.4 & \bf 69.9 $\pm$ 3.2 & \bf 81.9 $\pm$ 8.0 & 73.0 $\pm$ 4.2 & \bf 60.8 $\pm$ 6.1 &  76.1 $\pm$ 2.3 & \bf 66.5 & -\\
 
        \bottomrule
    \end{tabular}}
    \caption{Test accuracy on the TUDatasets with 10-fold cross-validation. Standard deviations are indicated with $\pm$. Column "\textsc{Avg}" shows the average test accuracy across data sets and column "p" indicates p-values of paired one-sided Wilcoxon tests across all datasets and validation folds comparing each method to the corresponding \textsc{vpa} variant.
}
    \label{tab:main_results}
    
\end{table}

\section{Discussion}
\label{sec:discussion}

Our results hint at a potentially powerful new aggregation function with equal expressivity as \textsc{sum} aggregation and improved learning dynamics. 

In general, it needs to be considered that better prediction performance of more powerful GNNs will only be observed when the underlying machine learning problem requires such a level of expressiveness. For benchmarks from real-world data, it might, however, not be known whether less powerful GNNs can also show competitive prediction performance.

Furthermore, variance preservation seems to be an important property to avoid exploding or vanishing activations. This is especially relevant for very deep networks. For the datasets used, all methods could be trained without diverging due to exploding activations. One reason could be that the GNNs are quite shallow and therefore there are only a few message-passing steps. Nevertheless, results in \Cref{tab:main_results} and learning curves in \Cref{fig:learning_curves} show that \textsc{vpa} has advantages over \textsc{sum} aggregation in terms of convergence speed.

On the social network datasets, \textsc{vpa} seems to perform particularly well compared to other methods when no additional node features are introduced, forcing the GNNs to learn from the network structure instead (see experimental results in \Cref{tab:main_results}). However, including the node degree as a feature improves the performance of less expressive GNNs (see \Cref{tab:results2}). The advantage in prediction performance of \textsc{vpa} over other methods is less pronounced in this setting.

While we suggest \textsc{vpa} as a general aggregation scheme, which is easily applicable to many GNN architectures, such as GIN, its application might not be obvious for other models. For example, \textsc{SGC} inherently contains a normalization strategy using node degrees and \textsc{GAT} makes use of attention weights during aggregation. In both cases, signal propagation is affected. Taking this into account, we suggest variants of \textsc{vpa} for \textsc{SGC} and \textsc{GAT}. Variance preservation for \textsc{GAT+vpa} is shown in \Cref{lemma:vpp_att}, however, we did not formally proof variance preservation for \textsc{SGC+vpa}.

It should further be considered, that distributional assumptions to formally show variance preservation might only hold at the time of initialization. However, as discussed in \Cref{sec:signal_prop_mpnn} even that time point is important. Furthermore, even under other assumptions on the distribution of the messages, arguments about the increase and decrease of variance would hold.

\begin{ack}
The ELLIS Unit Linz, the LIT AI Lab, the Institute for Machine Learning, are supported by the Federal State Upper Austria. We thank the projects Medical Cognitive Computing Center (MC3), INCONTROL-RL (FFG-881064), PRIMAL (FFG-873979), S3AI (FFG-872172), DL for GranularFlow (FFG-871302), EPILEPSIA (FFG-892171), AIRI FG 9-N (FWF-36284, FWF-36235), AI4GreenHeatingGrids (FFG- 899943), INTEGRATE (FFG-892418), ELISE (H2020-ICT-2019-3 ID: 951847), Stars4Waters (HORIZON-CL6-2021-CLIMATE-01-01). We thank NXAI GmbH, Audi.JKU Deep Learning Center, TGW LOGISTICS GROUP GMBH, Silicon Austria Labs (SAL), FILL Gesellschaft mbH, Anyline GmbH, Google, ZF Friedrichshafen AG, Robert Bosch GmbH, UCB Biopharma SRL, Merck Healthcare KGaA, Verbund AG, GLS (Univ. Waterloo), Software Competence Center Hagenberg GmbH, Borealis AG, T\"{U}V Austria, Frauscher Sensonic, TRUMPF and the NVIDIA Corporation.
\end{ack}

\clearpage

\small

\bibliography{refs}
\bibliographystyle{ml_institute}

\clearpage

\appendix

\counterwithin{figure}{section}
\counterwithin{table}{section}
\counterwithin{equation}{section}
\counterwithin{lemma}{section}
\renewcommand\thefigure{\thesection\arabic{figure}}
\renewcommand\thetable{\thesection\arabic{table}}
\renewcommand\theequation{\thesection\arabic{equation}}
\renewcommand\thelemma{\thesection\arabic{lemma}}

\section*{Appendix}

\section{Theoretical Details}

\subsection{Aggregation vs. Pooling} 
\label{sec:agg_vs_pool}

The message aggregation step and the graph-level readout step are critical operations in GNNs \citep{corso2020principal}. Message passing on graphs involves the pair-wise exchange of messages, a message aggregation mechanism, which combines messages from all neighboring nodes into one representation, and subsequent updates on nodes. This process can be linked to convolution operations~\citep{wu2020comprehensive, kipf2016semi, bronstein2021geometric}. However, unlike traditional convolutions, where the kernel size remains fixed, the message aggregation in GNNs is contingent on the number of neighboring nodes and, consequently, the incoming messages ~\citep{wu2020comprehensive}. 

For graph-level readouts, the distributed neural representation across the graph needs to be fused to a common representation space. This operation is denoted as pooling in the context of GNNs. For CNNs pooling often also refers to the aggregation step itself. We will however be more strict in distinguishing aggregation and pooling here and consider pooling to be caused by the stride parameter of CNNs. Graph-level readout pooling operations can be grouped into topology-based pooling, hierarchical pooling, and global pooling~\citep{lee2019self}. 
Global pooling consolidates the graph information into a single hidden representation before making final predictions, so similar operations as for message aggregation can be used here. Advanced pooling mechanisms consider the graph as a distribution, from which nodes are sampled \citep{Chen2023}.

\subsection{MLP Signal Propagation} 
\label{sec:signal_prop_mpnn}

In accordance with signal propagation literature \citep{schoenholz2016deep, klambauer2017self} we are interested in signal propagation of randomly initialized neural networks, i.e., we assume distributions on weights of these networks. Although it might also seem interesting to know about signal propagation behavior at different time points during training, this is much more difficult to study, since the distributions of weights might then also depend on the training data. However, an argument for studying signal propagation at initialization would be that learning might not work at all (or start well) when signal propagation throughout the whole network does not even work (well) at initialization.

In order to investigate the forward dynamics of a message-passing
network at initialization time with signal propagation theory, 
we take the following assumptions, assuming the case of $\phi$ taking two arguments\footnote{For the one-argument version of $\phi$ (where the message is computed only from the node representation $\Bh_j$ of the neighboring node) the line of reasoning is almost analogous.}. The initial representation
of pairs of node representations $\Bh^P_{ij}=\left(\Bh_i,\Bh_j\right)$ with $i\neq j$ 
follows a data distribution
$\Bh^P_{ij} \sim p_{\mathrm{data}}$ 
with some mean
$\EXP_{\Bh^P \sim p_{\mathrm{data}}}\left( \Bh^P_{ij} \right) = \Bmu_{\Bh^P}$
and some covariance 
$\COV_{\Bh^P \sim p_{\mathrm{data}}}\left( \Bh^P_{ij} \right) = \boldsymbol C_{\Bh^P}$.

We further assume a deep and broad MLP $\phi_{\Bw}(.)$ 
with randomly sampled weights
according to LeCun's initialization scheme \citep{lecun1998efficient}, 
$w \sim p_{\mathcal N}(0,1/H)$, where $H$ is 
the fan-in of each neuron,
and with linear activation in the last layer. 
Since an MLP $\phi$ is a 
measurable function,
$\Bm_{ij}=\phi_{\Bw}(\Bh_i,\Bh_j)$
is also a random variable. 
Then, central results from 
signal propagation theory \citep{neal1995bayesian,schoenholz2016deep,lee2017deep,hoedt2023principled} imply that the distribution of $\Bm_{ij}$ at initialization
can be approximated by a standard normal distribution
$\Bm_{ij} \sim p_{\mathcal N}(\boldsymbol 0, \BI)$
\citep[Section 2.2]{lee2017deep}
and even a fixed point at zero-mean and unit variance
can be enforced \citep{klambauer2017self,lu2023bidirectionally}.
In practice, batch- \citep{Ioffe2015} or layer-norm \citep{Ba2016} are often used 
in these MLPs to partly maintain these statistics, i.e. zero mean
and unit variance, also during learning.
We are aware that this approximation only holds at initialization
and might be overly simplistic \citep{martens2021rapid} 
(see \Cref{sec:discussion}). However, note 
that we use this assumption only to make the 
point of variance preservation of the aggregation step. Even under 
other assumptions on the distribution of $\Bm_{ij}$ the arguments
about increase and decrease of variance would hold.

\subsection{Proof ~\Cref{lemma:vpp}}
\label{sec:var_pres}

\begin{proof}
    Because the variables $z_n$ are centered, we have
    \begin{align}
        \EXP[y]= \EXP \left[\frac{1}{\sqrt{N}}\sum_{n=1}^N z_n \right] = 
                 \frac{1}{\sqrt{N}}\sum_{n=1}^N \EXP[z_n]=0=\EXP[z].
    \end{align}

    Furthermore, we have 
    \begin{align}
        \VAR[y]&= \EXP\left[ \left(\frac{1}{\sqrt{N}}\sum_{n=1}^N z_n \right)^2\right] - \EXP\left[\frac{1}{\sqrt{N}}\sum_{n=1}^N z_n\right]^2 = \\
                &=\EXP \left[ \frac{1}{N} \left(\sum_{n=1}^N z_n \right)^2\right] =
                \frac{1}{N}\EXP \left[\sum_{n=1}^N z_n^2+     \sum_{n=1}^N \sum_{m=1,m \neq n}^N 2 z_n z_m \right] = \\
                &= \frac{1}{N} N \EXP[z_n^2] = \VAR[z_n] = \VAR[z],
    \end{align}

    where we have used the independence assumption $ \EXP[z_n z_m]=\EXP[z_n]\EXP[z_m]=0$ and that the $z_n$ are centered, which means that $\EXP[z_n^2]= \VAR[z_n]$.

\end{proof}

\subsection{Proof ~\Cref{lemma:expressivity}}
\label{sec:expr}

\begin{proof}
    Since the number of unique elements in $\mathcal{X}$ is bounded by $N$, there exists a bijective mapping $Z: \mathcal{X} \rightarrow \{1, ..., N\}$ assigning a natural number to each $x \in \mathcal{X}$. Then an example of such a function $f$ is a one-hot encoding function $f(x) = e_{Z(x)}$, with $e_{Z(x)} \in \mathbb{R}^N$ being a standard basis vector, where component $i$ of $e_{Z(x)}$, i.e. $e_{Z(x)}[i]$ is defined as $e_{Z(x)}[i]:=\begin{cases}
0 & \text{for}\quad i \neq Z(x)\\
1 & \text{for}\quad i = Z(x)
\end{cases}$.

We define $h(X)$ to be:
$$h(X) = \frac{1}{\sqrt{|X|}} \sum_{x \in X} f(x) = \frac{1}{\sqrt{|X|}} \sum_{x \in X} e_{Z(x)}.$$\\

Summing up the components of $h(X)$ yields the square root of the cardinality of $X$, i.e. the embeddings contain information on the cardinality of $X$. Since we know, $\sqrt{|X|}$ from the embedding, we can just multiply the embedding $h(X)$ with $\sqrt{ |X|}$ to obtain the original multiplicity of each element $x$ in multiset $X$. Thus, the multiset $X \subset \mathcal{X}$ can be uniquely reconstructed from $h(X)$, implying that $h$ is injective.

\end{proof}

We note, that for \textsc{mean} aggregation, i.e., $\tilde h(X) = \frac{1}{|X|} \sum_{x \in X} f(x)$, the multiset $X$ cannot be reconstructed from $\tilde h(X)$, since in that case the components of $\tilde h(X)$ sum up to 1 and therefore, do not indicate the cardinality of $X$ (e.g., $h(\{0, 1\}) = (0.5,0.5) = h(\{0, 0, 1, 1\})$). In contrast, for $h(X) = \frac{1}{\sqrt{|X|}} \sum_{x \in X} f(x)$,  the embeddings contain information on the cardinality of $X$, which is lost for \textsc{mean} aggregation. Multiplication by $|X|$ does not work for \textsc{mean} aggregation to reconstruct the original multiset $X$, as no cardinality information is stored in the embedding $\tilde h(X)$. More generally, no function $f$ can be found such that $\tilde h(X)$ is unique for each multiset $X \subset \mathcal{X}$ of bounded size (see Corollary 8 in \citet{Xu2019}).

\subsection{Proof ~\Cref{lemma:vpp_att}}
\label{sec:proof_gat}

\begin{proof}
    Because the variables $z_n$ are centered, we have
    \begin{align}
        \EXP[y]= \EXP \left[\frac{1}{C}\sum_{n=1}^N c_n z_n \right] = 
                 \frac{1}{C}\sum_{n=1}^N c_n \EXP[z_n]=0=\EXP[z].
    \end{align}

    Furthermore, we have 
    \begin{align}
        \VAR[y]&= \EXP\left[ \left(\frac{1}{C}\sum_{n=1}^N c_n z_n \right)^2\right] - \EXP\left[\frac{1}{C}\sum_{n=1}^N c_n z_n\right]^2 = \\
                &=\EXP \left[ \frac{1}{C^2} \left(\sum_{n=1}^N c_n z_n \right)^2\right] =
                \frac{1}{C}\EXP \left[\sum_{n=1}^N c_n^2 z_n^2+     \sum_{n=1}^N \sum_{m=1,m \neq n}^N 2 c_n c_m z_n z_m \right] = \\
                &= \frac{1}{C^2} \sum_{n=1}^N c_n^2 \EXP[z_n^2]
                = \frac{1}{\sum_{i=1}^N c_i^2} \Big(\sum_{i=1}^N c_i^2\Big) \EXP[z_n^2] = \VAR[z_n] = \VAR[z],
    \end{align}

    where we have used the independence assumption $ \EXP[z_n z_m]=\EXP[z_n]\EXP[z_m]=0$ and that the $z_n$ are centered, which means that $\EXP[z_n^2]= \VAR[z_n]$.

\end{proof}

Note, that for case of uniform attention weights, $C=\sqrt{\sum_{i=1}^N c_i^2}=\sqrt{\sum_{i=1}^N \Big(\frac{1}{N}\Big)^2}=\\ \sqrt{N \frac{1}{N^2}}=\frac{1}{\sqrt{N}}$. Further $y=\frac{1}{C} \ \sum_{i=1}^N c_i \ z_i=\frac{1}{\frac{1}{\sqrt{N}}} \ \sum_{i=1}^N \frac{1}{N} \ z_i=\sqrt{N} \ \frac{1}{N} \sum_{i=1}^N  \ z_i=\frac{1}{\sqrt{N}} \sum_{i=1}^N  \ z_i$ is obtained, which is the same as \textsc{vpa}.\\
In the case that attention focuses on exactly one value, i.e., $c_j=1$ and $c_i=0 \ \ \forall \ i \neq j$ gives $C=1$, and $y=\frac{1}{C} \ \sum_{i=1}^N c_i \ z_i=z_j$. Cardinality information is lost in this case. However, the attention mechanism might be learnable and therefore not converge to this solution if limited expressivity leads to larger losses during optimization.

\clearpage

\section{Experimental Details \& Further Results}

\subsection{Implementation Details}
\label{sec:hyperparams}

We extended our framework upon implementations as provided by PyTorch Geometric (\cite{Fey2019}). Specifically, we used the following convolutional layers: GINConv (\textsc{GIN}), GraphConv (\textsc{GCN}), SGConv (\textsc{SGC}) and GATConv (\textsc{GAT}).
We used $5$ GNN layers for \textsc{GIN}, \textsc{GCN} and \textsc{GAT}, respectively, and one layer with $K=5$ hops for SGC. The dimension of the messages was $64$ for all architectures. An MLP with one hidden layer was used for classification with a hidden dimension of $64$ for \textsc{GIN} and
$128$ for all other models. Furthermore, we used a dropout rate of $0.5$ and the standard Adam optimizer with a learning rate of $0.001$.

\subsection{Extended results}

\Cref{tab:results2} shows results for the social datasets in the TUDataset benchmark with the node degree encoded as node features. Please refer to \Cref{sec:experiments} for further details and to \Cref{sec:discussion} for a discussion of these results compared to those in \Cref{tab:main_results}.

\begin{table}[ht]
    \centering
    \resizebox{0.7\textwidth}{!}{
    \begin{tabular}{lcccccc}
        \toprule 
        &  {\textsc{IMDB-B}} & {\textsc{IMDB-M}}  & {\textsc{RDT-B}} & {\textsc{RDT-M5K}} & {\textsc{COLLAB}} \\
        \midrule
        {\textsc{GIN+sum}} & 72.5  $\pm$ 4.5 & \bf 50.8  $\pm$ 4.1 & 81.5  $\pm$ 1.7 & \bf 47.5  $\pm$ 2.4 & \bf 82.2  $\pm$ 1.7 
        \\ 
        {\textsc{GIN+mean}} & \bf 73.8  $\pm$ 4.4 & 48.9  $\pm$ 3.7 & 77.1  $\pm$ 2.8 & 47.1  $\pm$ 1.6 & 80.7  $\pm$ 1.0 
        \\ 
        {\textsc{GIN+max}} & 71.0  $\pm$ 4.5 & 47.5  $\pm$ 4.9 & 78.5  $\pm$ 2.2 & 42.7  $\pm$ 2.1 & 77.1  $\pm$ 1.7 
        \\ 
        {\textsc{GIN+vpa}} & 73.7  $\pm$ 3.7 & 49.7  $\pm$ 3.6 & \bf 82.0  $\pm$ 2.0 & 47.4  $\pm$ 1.9 & \bf 82.2  $\pm$ 1.7 
        \\ 
        \midrule
        {\textsc{GCN+sum}} & 70.7 $\pm$ 3.1 & 43.9 $\pm$ 3.7 & 76.3 $\pm$ 3.6 & \bf 50.4 $\pm$ 2.4 & 73.7 $\pm$ 2.2\\
        {\textsc{GCN+mean}} & 71.9 $\pm$ 5.2 & \bf 51.3 $\pm$ 3.4 & 71.0 $\pm$ 2.5 & 46.3 $\pm$ 2.3 & 80.6 $\pm$ 1.0\\
        {\textsc{GCN+max}} & 62.9 $\pm$ 3.5 & 43.1 $\pm$ 4.2 & 63.4 $\pm$ 5.0 & 30.6 $\pm$ 2.6 & 74.8 $\pm$ 1.6\\
        {\textsc{GCN+vpa}} & \bf 73.6 $\pm$ 5.5 & 50.5 $\pm$ 2.7 & \bf 80.6 $\pm$ 3.4 & 47.9 $\pm$ 2.3 & \bf 81.3 $\pm$ 1.5\\

        \midrule
        {\textsc{SGC}} & \bf 72.9 $\pm$ 3.9 & \bf 50.6 $\pm$ 3.5 & 81.0 $\pm$ 2.4 & \bf 49.0 $\pm$ 1.9 & \bf 81.3 $\pm$ 1.8\\
        {\textsc{SGC+vpa}} & 72.6 $\pm$ 3.7 & 49.4 $\pm$ 3.6 & \bf 81.5 $\pm$ 2.3 & 47.8 $\pm$ 2.8 & 80.5 $\pm$ 1.1\\
        
        \midrule
        {\textsc{GAT}} & \bf 73.9 $\pm$ 3.4 & \bf 50.2 $\pm$ 4.0 & 78.3 $\pm$ 3.0 & 47.0 $\pm$ 2.7 & \bf 81.2 $\pm$ 1.4\\
        {\textsc{GAT+vpa}} & 71.7 $\pm$ 4.9 & 49.6 $\pm$ 6.1 & \bf 79.1 $\pm$ 2.3 & \bf 47.5 $\pm$ 1.7 & 79.5 $\pm$ 1.5\\

        \bottomrule
    \end{tabular}}
    \caption{Results on the social datasets of the benchmark setting by \citep{Xu2019}. In this variant of the datasets, the number of neighbors of a node is encoded as a node feature. The compared methods are again \textsc{GIN} and \textsc{GCN} with four different aggregation functions and \textsc{SGC} and \textsc{GAT} with their tailor-made variance preservation modifications.}
    \label{tab:results2}
\end{table}

\clearpage

\subsection{Learning Dynamics}
\label{sec:learningCurves}
\begin{figure}[h]
    \centering
    \includegraphics[width=\textwidth]{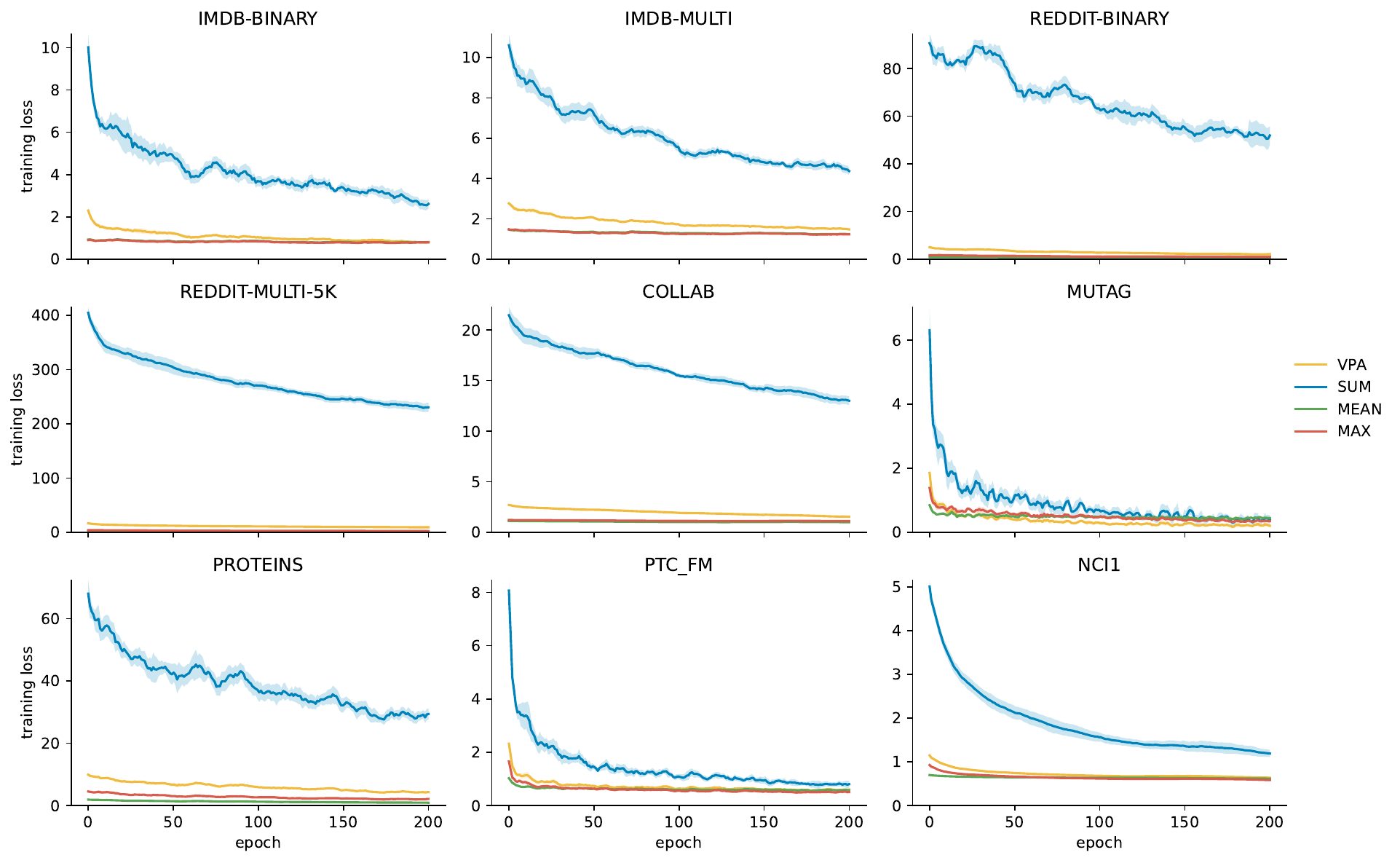}
    \caption{Learning Curves of the \textsc{GIN} architecture with different aggregation functions on the TUDataset benchmarks used by \citet{Xu2019} and which were retrieved in the version as provided by \citet{Morris2020}. Note that the default hyperparameters are adjusted to the \textsc{sum} aggregation function. Nevertheless, the network training converges faster with variance-preserving aggregation (\textsc{vpa}) compared to \textsc{sum} aggregation. At the same time, \textsc{vpa} also maintains expressivity, whereas \textsc{mean} and \textsc{max} aggregation decrease the expressivity of GNNs.}
    \label{fig:learning_curves}
\end{figure}

We investigated the learning dynamics of the compared methods
based on the training loss curves (see Figure~\ref{fig:learning_curves}).
The learning curves show that \textsc{GIN} model training converges 
fast with \textsc{mean}, \textsc{max} and \textsc{vpa} and slower
with \textsc{sum} aggregation, which we attribute to the exploding variance
in the forward pass.

\clearpage

\end{document}